\documentclass[a4paper,conference]{IEEEtran}
\usepackage{cite}

\ifCLASSINFOpdf
  \usepackage[pdftex]{graphicx}
\else
\fi
\usepackage{amsmath}
\usepackage{array}

\usepackage{multirow}

\ifCLASSOPTIONcompsoc
  \usepackage[caption=false,font=normalsize,labelfont=sf,textfont=sf]{subfig}
\else
  \usepackage[caption=false,font=footnotesize]{subfig}
\fi

\begin{document}
%
\title{Unsupervised Disentanglement GAN for\\ Domain Adaptive Person Re-Identification}

\author{\IEEEauthorblockN{Yacine Khraimeche\IEEEauthorrefmark{1}, Guillaume-Alexandre Bilodeau\IEEEauthorrefmark{1}, David Steele\IEEEauthorrefmark{2}, and Harshad Mahadik\IEEEauthorrefmark{2}}
\IEEEauthorblockA{\IEEEauthorrefmark{1}LITIV lab., Polytechnique Montréal, Montréal QC, H3T 1J4, Canada}
\IEEEauthorblockA{\IEEEauthorrefmark{2}Arcturus Networks, Etobicoke, ON, M9C 1A3, Canada}
\IEEEauthorblockA{\{yacine.khraimeche, gabilodeau\}@polymtl.ca \\
\{dsteele, harshad\}@arcturusnetworks.ca}}


%


\maketitle

\begin{abstract}
While recent person re-identification (ReID) methods achieve high accuracy in a supervised setting, their generalization to an unlabelled domain is still an open problem. In this paper, we introduce a novel unsupervised disentanglement generative adversarial network (UD-GAN) to address the domain adaptation issue of supervised person ReID. Our framework jointly trains a ReID network for discriminative features extraction in a source labelled domain using identity annotation, and adapts the ReID model to an unlabelled target domain by learning disentangled latent representations on the domain. Identity-unrelated features in the target domain are distilled from the latent features. As a result, the ReID features better encompass the identity of a person in the unsupervised domain. We conducted experiments on the Market1501, DukeMTMC and MSMT17 datasets. Results show that the unsupervised domain adaptation problem in ReID is very challenging. Nevertheless, our method shows improvement in half of the domain transfers and achieve state-of-the-art performance for one of them.
\end{abstract}


%
\IEEEpeerreviewmaketitle


\section{Introduction}
Person re-identification (ReID) aims at matching pedestrian identities across a non-overlapping video surveillance camera network. This fine-grained recognition task is a key component for several video analysis tasks in connected large-scale camera network in smart cities, such as multi-target multi-camera tracking or person retrieval. It is a challenging task due to the large intra-class variation caused by multiple factors, for example pose, viewpoint, background, lighting, occlusions, camera calibration, etc.

Despite these difficulties, deep learning models have recently improved significantly the performance of person ReID methods \cite{zhengPersonReidentificationPresent2016}, even surpassing human accuracy \cite{zhangAlignedReIDSurpassingHumanLevel2018}. A ReID network learns to generate discriminative features encapsulating the identity of a person while being independent to the intra-class variations factors. 

\begin{figure}[t]
\centering
\includegraphics[width=3.47in,trim={1.4cm 0 1.4cm 0},clip]{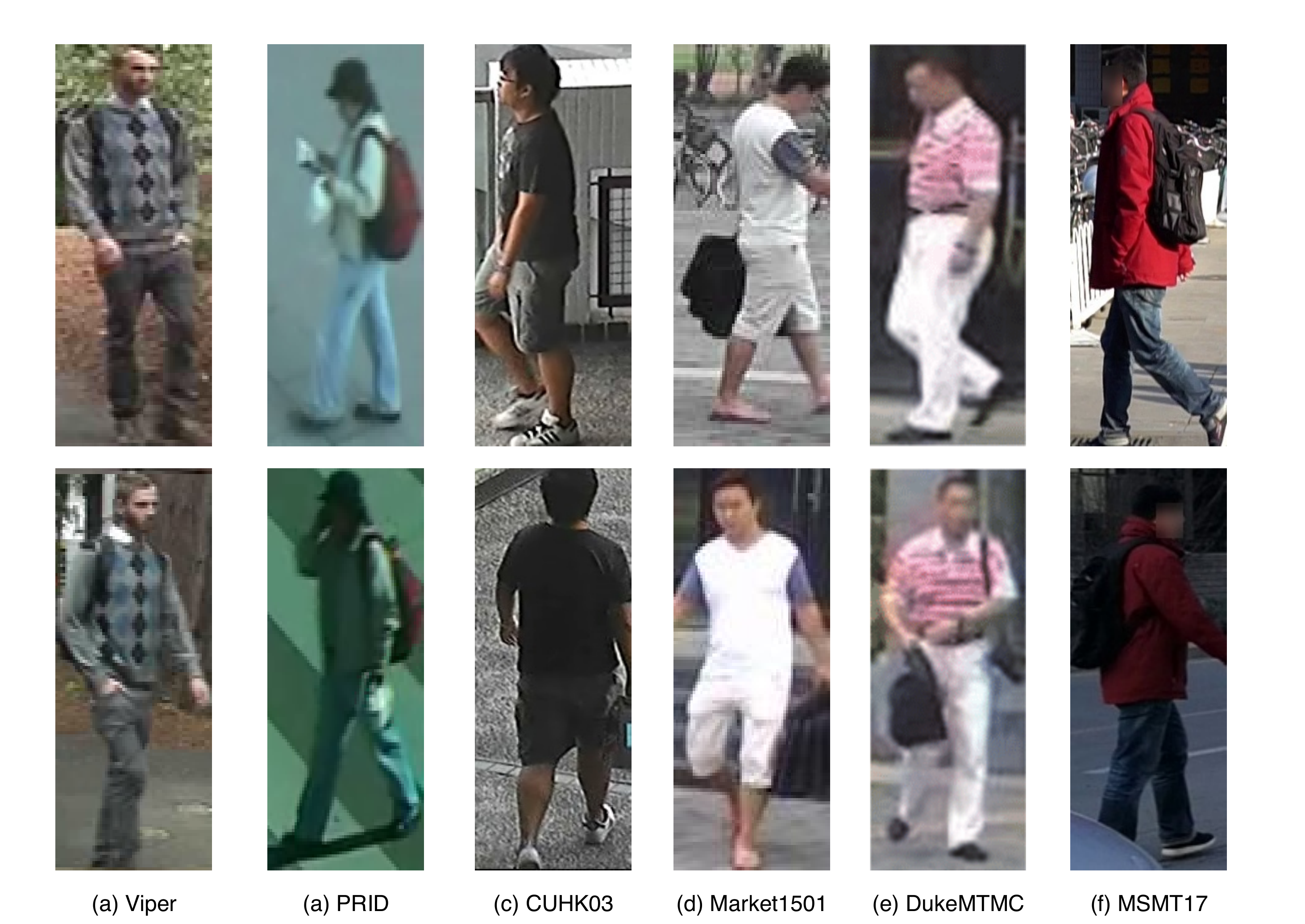}
\caption{Examples of image pairs from various ReID datasets : column a) Viper \cite{grayViewpointInvariantPedestrian2008a}, column b) PRID \cite{hirzerPersonReidentificationDescriptive2011}, column c) CUHK03 \cite{liDeepReIDDeepFilter2014}, column d) Market1501 \cite{zhengScalablePersonReidentification2015}, column e) DukeMTMC-reID \cite{ristaniPerformanceMeasuresData2016, zhengUnlabeledSamplesGenerated2017} and column f) MSMT17 \cite{weiPersonTransferGAN2018}. They showcase very different capture conditions with some including blur, lighting variations, low resolution, etc.}
\label{fig:DatabaseComparaison}
\end{figure}

However, this statement is only true in a supervised setting in which we have access to a training set of domain-related labelled images. ReID features do not generalize well. When the features are applied to a domain different from the one they were trained on, the performance of ReID models collapses since the output feature vectors are not adapted to this new domain. As shown in Figure~\ref{fig:DatabaseComparaison}, images from the new domain can be widely different from the training images because of different environment, lighting, viewpoint, camera settings and model, etc. All added variations are factoring with the intra-class ones, and the ReID network fails since it is not trained to deal with these.

Unfortunately, for real-world applications with large camera networks, it is impossible to collect and label large amount of data for each new use case/camera setup. The unsupervised setting, in which labelled images are not available, is a more realistic context for practical use of ReID. In this context, domain adaptive person ReID, that is to say optimizing a ReID network trained on a labelled source domain to an unlabelled target domain, is a promising research avenue to explore to enable large-scale real-world applications of ReID.

Some recent domain adaptive methods tackle the problem of unsupervised person ReID \cite{yuUnsupervisedPersonReIdentification2019, yangPatchBasedDiscriminativeFeature2019, zhouLearningGeneralisableOmniScale2019}. However, there is still room for improvement as the performance in the unsupervised setting is still far below its supervised counterpart.

\begin{figure*}
\centering
\includegraphics[width=5.5in]{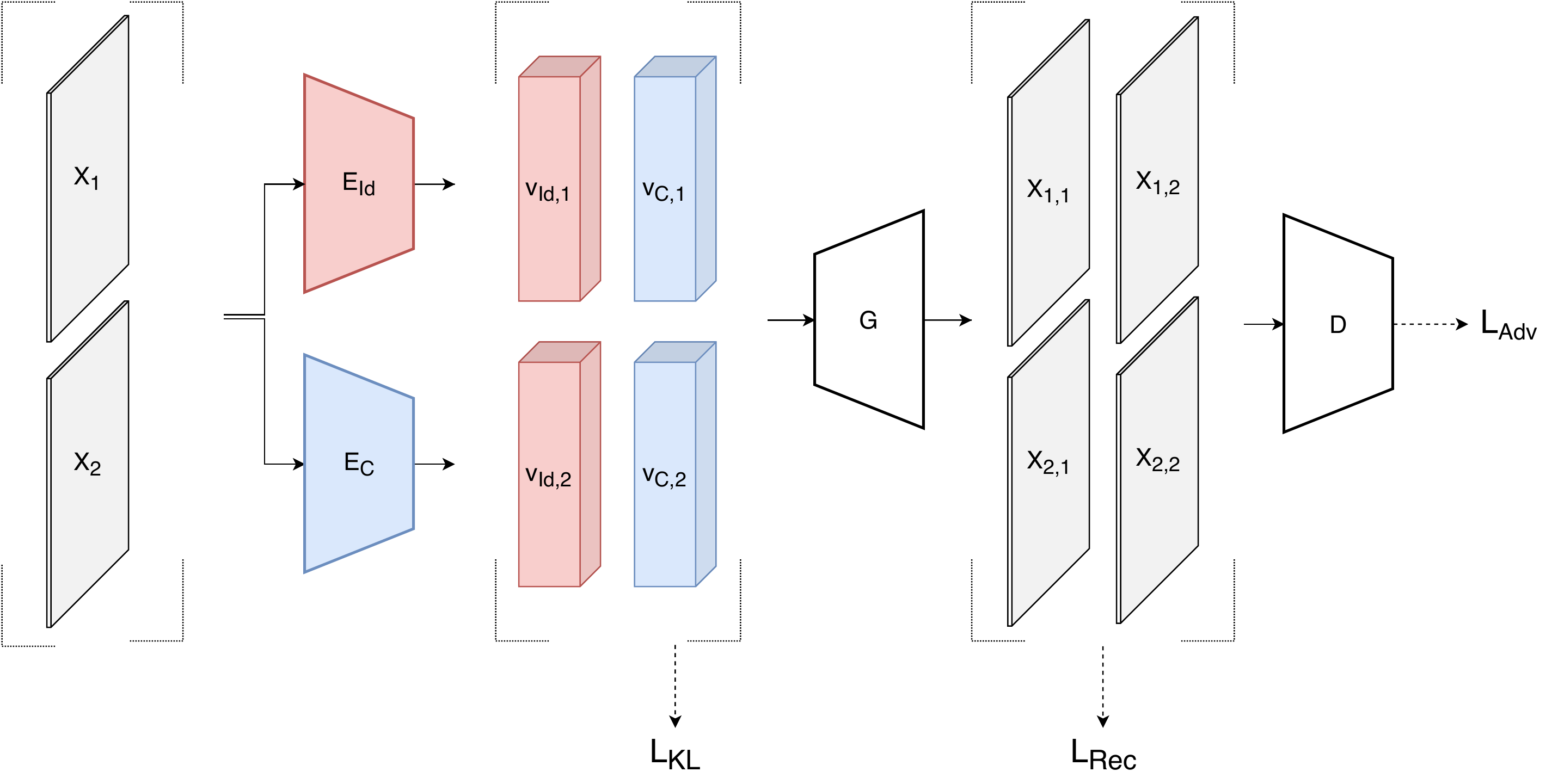}
\caption{UD-GAN framework overview. The identity encoder $E_{Id}$ and the content encoder $E_{C}$ extract the identity vectors $v_{Id}$ and the content vectors $v_{C}$, respectively, from the pair of images $X_{1}$ and $X_{2}$ of the target domain containing the same identity. The generator $G$ outputs four images by swapping the identity and content features in the generated images. The discriminator $D$ discriminates generated images from real images from the target domain.}
\label{fig:FrameworkOverview}
\end{figure*}

To address the unsupervised setting in person ReID, we introduce a novel neural network framework that incorporates the ReID network in a Generative Adversarial Network (GAN), and that jointly trains them with an auxiliary task : features disentanglement in the unlabelled domain. The ReID network learns to extract discriminative features in the labelled source domain using identity annotations. However, due to the domain gap, these features do not encompass every factor of variation in the target domain. We operate in the latent feature space to distill identity-unrelated features from the ReID features. Previous work proved the effectiveness of disentanglement for supervised person ReID \cite{eomLearningDisentangledRepresentation2019}. Without any labelling, our method is able to enforce efficient constraints to encourage the ReID network to produce ReID vectors adapted for the new domain.

We summarize our contributions as follows : 
\begin{itemize}
    \item We present a novel unsupervised disentanglement GAN for domain adaptive person ReID. Previous methods only considered disentanglement for ReID in a supervised setting \cite{eomLearningDisentangledRepresentation2019}, or required pose \cite{liCrossDatasetPersonReIdentification2019}.
    \item Our framework can be used with any ReID architecture, and does not require additional computing during inference.
    \item We conducted experiments on standard ReID benchmarks, and report improvements in half of the domain transfer scenarios, with even state-of-the-art performance for domain adaptive person ReID in one case.
\end{itemize}


\section{Related Work}

Person ReID is an active research topic. Some authors focus on metric learning, using hard triplets mining \cite{hermansDefenseTripletLoss2017} or a classification loss \cite{zhaiDefenseClassificationLoss2018}. Efforts have also been made to extract robust and discriminative features using network architectures like ResNet \cite{heDeepResidualLearning2016}, originally designed for object classification benchmarks such as ImageNet \cite{dengImageNetLargescaleHierarchical2009}. Adjustments related to the specific task of person ReID include pose estimation \cite{zhaoSpindleNetPerson2017}. Part-based methods propose instead to slice the image horizontally \cite{wangLearningDiscriminativeFeatures2018}. Some methods use attention mechanisms \cite{liHarmoniousAttentionNetwork2018} to focus the learning process on the foreground while ignoring background. Finally, other researchers optimize their own architecture specifically for the ReID task \cite{zhouOmniScaleFeatureLearning2019a}, using a multi-scale design.

The Unsupervised Domain Adaptation (UDA) goal is to adapt a model trained on a source labelled dataset to an unlabelled target dataset. It can done by minimizing the Maximum Mean Discrepancy (MMD) \cite{grettonKernelMethodTwoSampleProblem2007} between the source and target domain to align the distributions in the two domains. However, in ReID, the classes in the source and target domain are completely distinct, thus requiring a specific method to leverage invariance properties of the ReID datasets \cite{zhongInvarianceMattersExemplar2019}. Instead of adapting the whole image, image patches can also be compared to get around the class discrepancy issue \cite{yuUnsupervisedPersonReIdentification2019}. An auxiliary dataset can also be leveraged for Soft Multilabel Learning \cite{yangPatchBasedDiscriminativeFeature2019}. Another take on this problem is to specifically design a ReID network with the generalization in mind, using Instance Normalization blocks. Without any further adaptation, the OSNet-AIN architecture \cite{zhouLearningGeneralisableOmniScale2019} achieves good results on unlabelled domains.

GANs \cite{goodfellowGenerativeAdversarialNets2014} have been applied in ReID, with two main objectives : 1) style transfer for cross-dataset person ReID and 2) data augmentation. The style transfer generative network CycleGAN \cite{zhuUnpairedImagetoImageTranslation2017} is applied in \cite{weiPersonTransferGAN2018} to close the gap between the source and target domains, by applying the style of the target domain on the source domain during training. In \cite{liuIdentityPreservingGenerative2019} the ReID network is trained on generated images in the style of each camera of the target domain, obtained using StarGAN \cite{choiStarGANUnifiedGenerative2018a} to perform image-to-image translation.

Disentangled representation is a way to explain latent features produced by generative networks, by separating independent factors of variations. It has seen recent applications in supervised person ReID : FD-GAN \cite{geFDGANPoseguidedFeature2018} attempt to disentangle pose from appearance using pose estimation. DG-Net \cite{zhengJointDiscriminativeGenerative2019} jointly learns to extract discriminative features and to generate images where appearance and structure are disentangled. ISGAN \cite{eomLearningDisentangledRepresentation2019} takes into account every possible factor of variation by disentangling identity-related and identity-unrelated features. In an unsupervised setting, PDA-Net \cite{liCrossDatasetPersonReIdentification2019} applies pose-guided disentanglement. In contrast, our method does not use pose annotations.


\section{Method}

\subsection{Framework overview}
Our proposed domain adaptation method is based on the ReID feature vectors disentanglement in the target domain. Following \cite{eomLearningDisentangledRepresentation2019}, we propose to disentangle identity features from every other identity-unrelated factors of variation of the latent features, that we name content features.  For this new task, the ReID network is incorporated in a generative framework \cite{goodfellowGenerativeAdversarialNets2014}. As the GAN learns to generate realistic images that preserve identity and content as an auxiliary task, the ReID network specializes in identity related features extraction in the target domain.

This auxiliary task does not require any identity annotation in the target dataset. It uses pairs of target domain images displaying the same identity that we extract using the ReID network pretrained on the source dataset. We propose a method to filter out wrong pairs to create a close to noise-free pair dataset in the target domain.

The generator takes features extracted from a pair of images as input. It learns to generate realistic images that will preserve both identity and content. Constraints on both the latent feature space and on the generated images in the target domain induce the domain adaptation of the identity features used in the generative process.

The disentanglement process involves a generative module, a discriminative module and two encoding modules, as shown in Figure~\ref{fig:FrameworkOverview}.

\subsection{Identity encoder}

Our framework disentangle identity features from identity-unrelated ones. To this end, it includes two encoders that output the two disentangled latent feature space. The ReID network $E_{Id}$ is used as an identity-related feature encoder that extracts identity-related feature vectors $v_{Id}$. $E_{Id}$ only partially encodes the images since identity-unrelated features are absent from $v_{Id}$. 

$E_{Id}$ is trained for person ReID on the labelled source domain. On this dataset, supervised person ReID is a multi-class classification task : each identity corresponds to a class. For this task, we add a fully connected classification layer to the ReID network and train it to associate the ReID vectors to the correct class using cross-entropy loss \cite{zhaiDefenseClassificationLoss2018}, and label smoothing. Thus, we define the identity loss $\mathcal{L}_{Id}$ on the source domain.



\subsection{Content encoder}

\begin{figure*}[ht]
\centering
\includegraphics[width=5in]{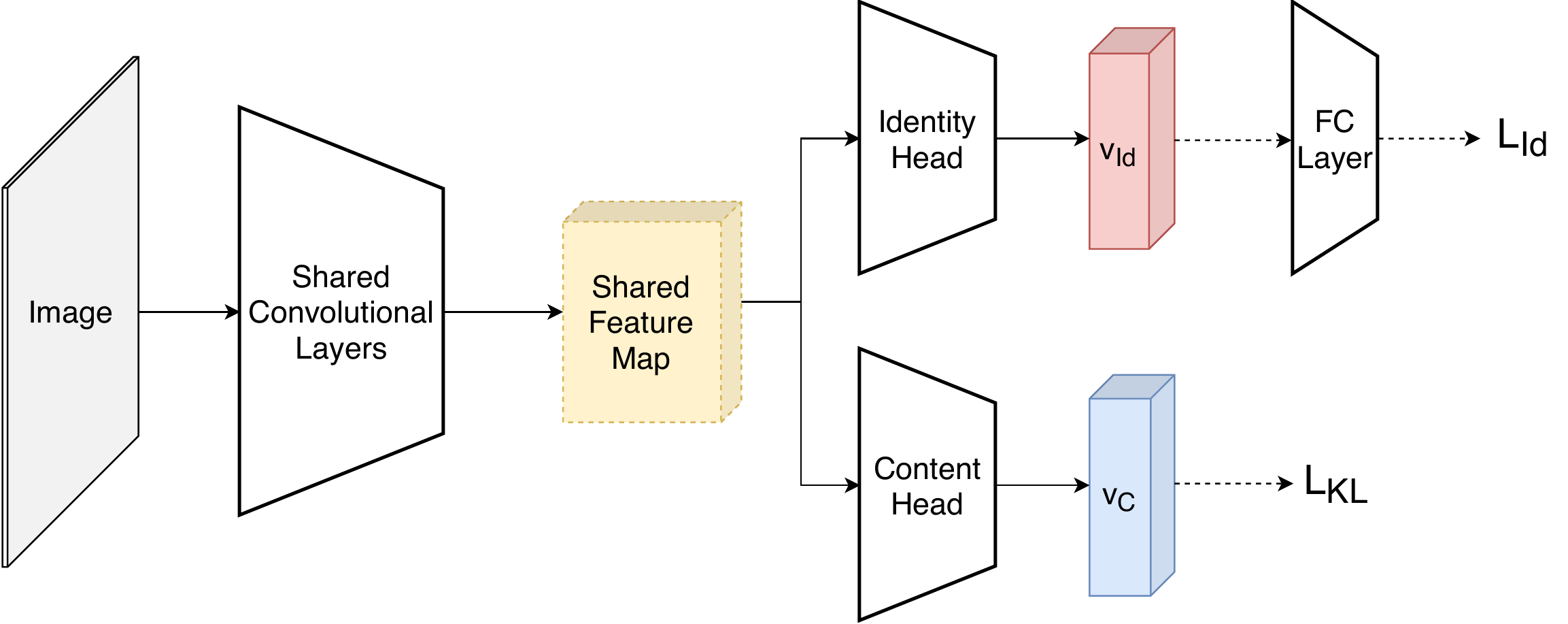}
\caption{Identity and Content Encoders architecture. The ReID network is composed of the shared convolutional layers and the identity head, with an added fully connected classification layer to compute $\mathcal{L}_{Id}$.}
\label{fig:EncodersArchitecture}
\end{figure*}

We introduce a second encoder to encompass identity-unrelated features, the content encoder $E_{C}$. As proposed in \cite{eomLearningDisentangledRepresentation2019}, $E_{C}$ is designed as a second head added to the ReID network (Figure~\ref{fig:EncodersArchitecture}). The first convolutional layers are shared between the two networks. The output of these layers is a common feature map for both encoders. The last layers of $E_{Id}$ are duplicated and trained for identity-unrelated features extraction. The content feature vector $v_{C}$ generated by the content encoder $E_{C}$ complements the information of $v_{Id}$, allowing for a complete latent representation of the original image from the features extracted by both encoders. Thus, we build a second latent feature space orthogonal to the identity-related features space.

Contrary to $E_{Id}$, $E_{C}$ is trained directly on the target domain by minimizing the common information shared between the vectors $v_{C}$ and $v_{Id}$. Without additional constraint on the content encoder $E_{C}$, we observe that the generator tends to use only the content vector information to generate images. Indeed, unlike $E_{Id}$ whose training is closely constrained by the identity loss $\mathcal{L}_{Id}$, the content encoder is free to extract any information from the images into $v_{C}$. That is why we introduce a regularization loss on the encoder $E_{C}$ to limit the amount of information stored in $v_{C}$, the Kullback-Leibler (KL) divergence loss :
\begin{equation}
    \mathcal{L}_{KL} = D_{KL}(v_{C}\parallel\mathcal{N}(0, 1)), 
\end{equation}
where 
\begin{equation}
    D_{KL}(p\parallel q)= - \int _{-\infty }^{\infty }p(x)\log \left({\frac {p(x)}{q(x)}}\right)\,dx.
\end{equation}

The KL divergence loss provides an incentive for both encoders to produce complementary information, thus ensuring the disentanglement process. We implement $\mathcal{L}_{KL}$ using the re-parameterization trick \cite{kingmaAutoEncodingVariationalBayes2014, luUnsupervisedDomainSpecificDeblurring2019}.

\begin{figure*}[t]
\centering
\includegraphics[width=3.8in]{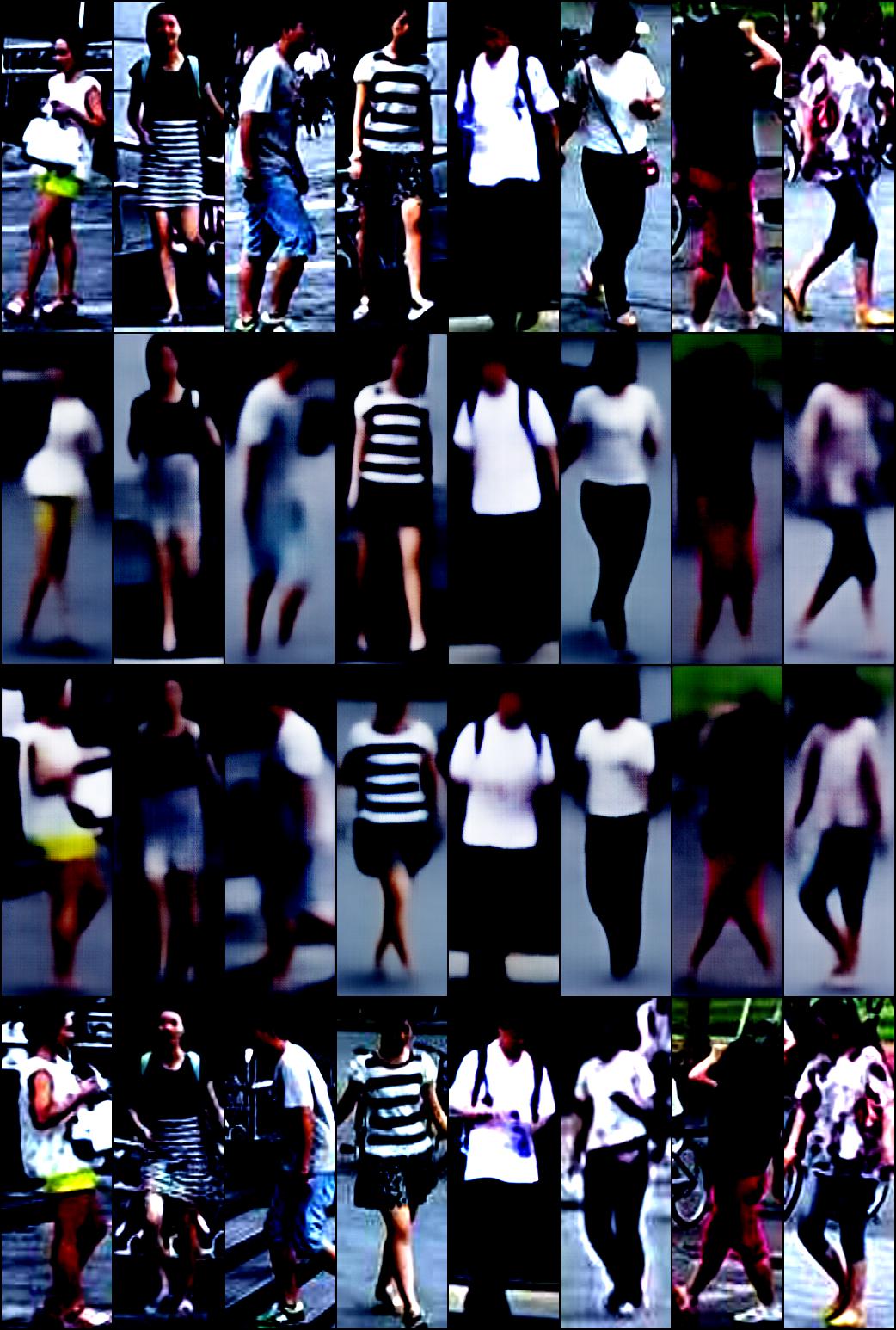}
\caption{Example of generated images during the transfer from MSMT17 to Market1501. First and last row : original image pairs from the Market1501 dataset. Second row : reconstructed images using identity and content features extracted from the same image in the first row. Third row : images generated by swapping identity from the first row and content from the last row.}
\label{fig:GeneratedImages}
\end{figure*}

\subsection{Generator and Discriminator}

Given two images $X_{1}$ and $X_{2}$ of the target domain containing the same identity, we extract two ReID vectors $v_{Id, 1}$ and $v_{Id, 2}$ and two content vectors $v_{C, 1}$ and $v_{C, 2}$.

The generator $G$ outputs four images from the pair of images by swapping the identity and content features in the generated images (Figure~\ref{fig:FrameworkOverview}). For example, $X_{1,2}$ will contains the same identity as $X_{1}$, while displaying the content of $X_{2}$.

It competes in an adversarial scheme with the discriminator $D$, that ensures generated images are realistic in the new domain. The discriminator extracts a feature map from original and generated images. We apply PatchGAN \cite{isolaImagetoImageTranslationConditional2017a} on the feature map to discriminate real images from generated ones. It operates at the scale of patches by applying a convolution, and averages the local outputs to produce a global score. The resulting adversarial loss $\mathcal{L}_{Adv}$ affects both the encoders and the generator :

\begin{equation}
     \mathcal{L}_{Adv} = \sum_{i\in \{1,2\}}^{} \log D(X_{i}) + \sum_{i,j \in \{1,2\}}^{} \log(1-D(G(v_{Id,i}, v_{C,j})).
\end{equation}

Image generation is also constrained using a reconstruction loss function. We define the reconstruction loss $\mathcal{L}_{Rec}$ on the encoders and the generator as :
\begin{equation}
    \mathcal{L}_{Rec} = \sum_{i,j \in \{1,2\}}^{} \lVert X_{i,j} - X_{i}\rVert_{1}.
\end{equation}

When $i = j$, the generated image $X_{i,i}$ and the original image $X_i$ should be identical : we want to ensure the information extracted by the two encoders is exhaustive. In the same way, in the case $i \neq j$, the reconstructed image $X_{i,j}$ should then be similar to the image $X_{j}$ where $v_{C, j}$ came from, since the ReID network would encode the same identity information for both $X_{i}$ and $X_{j}$.

\subsection{Loss functions}

We have defined two joint training objectives, one for each domain. On the source labelled domain, the reID network is trained for discriminative identity features extraction using $\mathcal{L}_{Id}$.

On the unlabelled target domain, the generative framework is trained for disentangled feature extraction and image generation, by optimizing the total loss on the target domain defined by :
\begin{equation}
     \mathcal{L}_{Target} = \lambda_{Rec}\mathcal{L}_{Rec} + \lambda_{KL}\mathcal{L}_{KL} + \lambda_{Adv}\mathcal{L}_{Adv}.
\end{equation}

The weights $\lambda_{Rec}$,  $\lambda_{KL}$ and $\lambda_{Adv}$ are hyperparameters of our framework.

\begin{table*}[htb]
\centering
\begin{tabular}{c|c|c|c|c|c|c|c|c} 
    \hline
    \multirow{2}{*}{Méthode} & \multicolumn{4}{c|}{MSMT17 $\,\to\,$ Market1501} & \multicolumn{4}{c}{MSMT17 $\,\to\,$ DukeMTMC} \\
    \cline{2-9}
    & R1 & R5 & R10 & mAP & R1 & R5 & R10 & mAP \\
    \hline
    MAR \cite{yuUnsupervisedPersonReIdentification2019} & 67.7 & 81.9 & - & 40.0 & 67.1 & 79.8 & - & 48.0 \\
    PAUL \cite{yangPatchBasedDiscriminativeFeature2019} & 68.5 & 82.4 &  87.4 & 40.1 & \textbf{72.0} & \underline{82.7} & \underline{86.0} & \textbf{53.2} \\
    OSNet-AIN \cite{zhouLearningGeneralisableOmniScale2019} & \underline{70.1} & \underline{84.1} & \underline{88.6} & \underline{43.3} & \underline{71.1} & \textbf{83.3} & \textbf{86.4} & \underline{52.7} \\
    \hline
    UD-GAN (Ours) & \textbf{73.6} & \textbf{86.4} & \textbf{90.7} & \textbf{47.2} & 62.7 & 75.6 & 80.3 & 43.6 \\
    \hline
\end{tabular}
\caption{Domain adaptation results from MSMT17 to Market1501 and DukeMTMC datasets. Bold indicates the best results, underscored the second best.}
\label{tab:ResultsMSMT17}
\end{table*}

\begin{table*}[htb]
\centering
\begin{tabular}{c|c|c|c|c|c|c|c|c} 
    \hline
    \multirow{2}{*}{Méthode} & \multicolumn{4}{c|}{DukeMTMC $\,\to\,$ Market1501} & \multicolumn{4}{c}{Market1501 $\,\to\,$ DukeMTMC} \\
    \cline{2-9}
    & R1 & R5 & R10 & mAP & R1 & R5 & R10 & mAP \\
    \hline
    IPGAN \cite{liuIdentityPreservingGenerative2019} & 57.2 & 76.0 & 82.7 & 28.0 & 47.0 & 63.0 & 68.1 & 27.0 \\
    ECN \cite{zhongInvarianceMattersExemplar2019} & \underline{75.1} & \textbf{87.6} & \textbf{91.6} & \underline{43.0} & \textbf{63.3} & \underline{75.8} & \underline{80.4} & \underline{40.4} \\
    PDA-Net \cite{liCrossDatasetPersonReIdentification2019} & \textbf{75.2} & \underline{86.3} & \underline{90.2} & \textbf{47.6} & \underline{63.2} & \textbf{77.0} & \textbf{82.5} & \textbf{45.1} \\
    OSNet-AIN \cite{zhouLearningGeneralisableOmniScale2019} & 61.0 & 77.0 & 82.5 & 30.6 & 52.4 & 66.1 & 71.2 & 30.5 \\
    \hline
    UD-GAN (Ours) & 61.6 & 77.2 & 83.0 & 31.4 & 49.7 & 64.7 & 70.5 & 30.2 \\
    \hline
\end{tabular}
\caption{Domain adaptation results between Market1501 and DukeMTMC datasets. Bold indicates the best results, underscored the second best.}
\label{tab:ResultsMarketDuke}
\end{table*}

\subsection{Training scheme}

\subsubsection{Stage 1 : ReID baseline pretraining}
The ReID network $E_{Id}$ is first trained using the identity loss $\mathcal{L}_{Id}$ on the labelled source domain to extract discriminative features while ignoring identity-independent elements such as pose, background or lighting. We adopted the OSNet-AIN \cite{zhouLearningGeneralisableOmniScale2019} architecture, a variant of OSNet \cite{zhouOmniScaleFeatureLearning2019a} optimized for domain adaptation. This design choice ensures the ReID network will perform reasonably well on the target domain from the end of the pretraining stage on the source domain, without having been exposed to the target domain.

\subsubsection{Stage 2 : Generative modules pretraining}

For the generative process, we need pairs of images of the same identity in the target domain. Since our method is unsupervised, we do not use any labels. To this end, we apply the pretrained ReID network on images of the target domain (the training set of the target dataset, ignoring the labels, in our experiments). We select the best matches according to the ReID distance for each image to create a pair. In order to reduce noise, we introduce a criteria to filter out possibly wrong matches. We establish for each image a ranked list of matches, according to the ReID distance. If the query image does not appear in the five best matches of its first ranking candidate, we filter out this pair. That is, both images of the pair should select each other based on the ReID distance. To have more data, discarded query images are included in pairs with twice the same image, replicating a situation where the two images of the pair are very similar. This post-processing step allows for a less noisy pair dataset to conduct the disentangling training on the target domain.

We freeze the identity encoder $E_{Id}$ and train only the content encoder $E_c$, the generator $G$ and the discriminator $D$ on these new pairs of images, with the target loss $\mathcal{L}_{Target}$.

\subsubsection{Stage 3 : Global training}
Finally, all modules are jointly trained on both domains, using $\mathcal{L}_{Id}$ on the source domain and $\mathcal{L}_{Target}$ on the target domain. Batches are alternatively sampled from the two domains. Here, the source dataset acts as a regularization to ensure that the ReID network $E_{Id}$ does not lose discriminative properties, while optimizing for image generation in the target domain. 


\section{Experiments}

\subsection{Implementation details}


\subsubsection{Architecture}

We use the OSNet-AIN \cite{zhouLearningGeneralisableOmniScale2019} architecture for the identity encoder $E_{Id}$. We add a fully connected layer when training on the source domain with cross-entropy loss. Five convolutional blocks compose the network, we duplicate the last two blocks and the final global average pooling layer to build the content encoder head $E_{C}$. We add two fully connected layers to the content head, to compute the re-parameterization trick for the KL divergence loss \cite{kingmaAutoEncodingVariationalBayes2014, luUnsupervisedDomainSpecificDeblurring2019}.

The Generator and Discriminator are built following \cite{eomLearningDisentangledRepresentation2019}. The Generator is composed of six blocks, each one containing a transposed convolutional layer \cite{radfordUnsupervisedRepresentationLearning2016a}, followed by batch normalization \cite{ioffeBatchNormalizationAccelerating2015a}, Leaky ReLU \cite{maasRectifierNonlinearitiesImprove2013b} and Dropout \cite{JMLR:v15:srivastava14a}. The Discriminator is a series of seven blocks composed of a convolutional layer followed by instance normalization \cite{ulyanovInstanceNormalizationMissing2017} and Leaky ReLU \cite{maasRectifierNonlinearitiesImprove2013b}. We then apply PatchGAN \cite{isolaImagetoImageTranslationConditional2017a} on the output feature map.

\subsubsection{Training} We resize the images to 384*128 and fix the content and identity feature vector sizes to 512. Following \cite{zhouLearningGeneralisableOmniScale2019}, the identity encoder is pretrained on ImageNet \cite{dengImageNetLargescaleHierarchical2009}. During stage 1, the identity encoder is fine-tuned on the whole source domain for 100 epochs with a batch size of 32. We use the AMSGrad \cite{reddiConvergenceAdam2018} optimizer with a initial learning rate of 0.00015, decayed using cosine annealing \cite{loshchilovSGDRStochasticGradient2017}. During the first 20 epochs, we only train the last classifier layer added to the pretrained network \cite{chenDeepTransferLearning2018}.

During stage 2, we train the discriminator using Stochastic gradient decent (momentum $=0.9$), the generator and the content encoder with the Adam \cite{kingmaAdamMethodStochastic2014} optimizer ($\beta_{1}=0.9, \beta_{1}=0.999$). The networks are trained for 200 epochs with a batch size of 16 images (8 pairs). We fix the learning rate to 2e-4.

On stage 3, we decay the learning rate to 2e-5 and train the whole framework for 400 epochs, with alternating steps on each domain.

\subsubsection{Hyperparameters} We performed a grid search to set the weights $\lambda_{KL} = 1e-4$, $\lambda_{Rec} = 10$ and $\lambda_{Adv} = 1$ in $\mathcal{L}_{Target}$. Specifically, the values of $\lambda_{KL}$ and $\lambda_{Rec}$ are a trade-off between achieving a better reconstruction by allowing more information in the content vector, and relying too much on the content vector, thus removing identity-related information from the identity vector.

We implemented our model using Pytorch, and trained it on a NVIDIA RTX2080 GPU.

\subsection{Datasets and evaluation}
We conducted experiments on the most recent and largest ReID datasets, Market1501 \cite{zhengScalablePersonReidentification2015}, DukeMTMC \cite{ristaniPerformanceMeasuresData2016, zhengUnlabeledSamplesGenerated2017} and MSMT17 \cite{weiPersonTransferGAN2018}. Detailed characteristics of these widely used datasets are summed up in Table~\ref{tab:CharDatasets}.

\begin{table}[ht]
\centering
\begin{tabular}{c|c|c|c} 
    \hline
    Dataset & Cam & Images & Identities (T-Q-G) \\
    \hline
    Market1501 & 6 & 32668 & 1501 (751-750-750)  \\
    DukeMTMC & 8 & 36411 & 1812 (702-702-1110) \\
    MSMT17 & 15 & 126441 & 4101 (1041-3060-3060) \\
    \hline
\end{tabular}
\caption{Market1501, DukeMTMC and MSMT17 dataset characteristics. (T: Train, Q: Query, G: Gallery), Cam: number of cameras}
\label{tab:CharDatasets}
\end{table}

We evaluate the domain adaptation from MSMT17 to Market1501 and DukeMTMC, and the domain adaptation between Market1501 and DukeMTMC. When used as the source dataset, the whole MSMT17 dataset is used for training, following the standard protocol for unsupervised domain adaptation \cite{zhouLearningGeneralisableOmniScale2019, yuUnsupervisedPersonReIdentification2019, yangPatchBasedDiscriminativeFeature2019}. Performance is measured using Cumulative matching characteristics (CMC) and mean average precision (mAP). The CMC curve represents, for a given rank k, the probability of finding the query identity among the first k results of the gallery. We report the rank 1, 5 and 10 accuracy. Mean average precision is defined as the mean area under the precision-recall curve.

\subsection{Results}

\subsubsection{Comparison with state-of-the-art}

We compared our method with recent state-of-the-art methods for domain adaptive person ReID. For the domain adaptation from MSMT17 : MAR \cite{yuUnsupervisedPersonReIdentification2019}, PAUL \cite{yangPatchBasedDiscriminativeFeature2019} and OSNet-AIN \cite{zhouLearningGeneralisableOmniScale2019}. Between Market1501 and DukeMTMC : IPGAN \cite{liuIdentityPreservingGenerative2019}, ECN \cite{zhongInvarianceMattersExemplar2019}, PDA-Net \cite{liCrossDatasetPersonReIdentification2019} and OSNet-AIN \cite{zhouLearningGeneralisableOmniScale2019}. We conducted our test under the same experimental settings and present the results in Table~\ref{tab:ResultsMSMT17} and Table~\ref{tab:ResultsMarketDuke}. 

Our method significantly outperforms previous methods for the transfer from MSMT17 to Market1501 : we report an improvement in absolute value of $3.5\%$ in Rank-1 accuracy and $3.9\%$ in mAP over the OSNet-AIN baseline on the Market1501 dataset. The disentanglement process identity features in the Market1501 target dataset, using the extracted pairs of images, has proven to be effective to improve the performance of the OSNet-AIN baseline. For this transfer, we exceed the performance of the state of the art.

However, our method does not improve the performance of the ReID reference network for the transfer from MSMT17 to DukeMTMC. On the contrary, the performance decreases in absolute value by $8.4\%$ in Rank-1 accuracy, and by $9.1\%$ in mAP compared to the baseline performance of OSNet-AIN on the DukeMTMC dataset.

We report the same results for the transfers between the DukeMTMC and Market1501 datasets. The domain adaptation process is effective to improve the performance of the baseline network when transferring from DukeMTMC to Market1501, since there is an increase in absolute value of $0.6\%$ in Rank-1 accuracy1, and $0.8\%$ in mAP. But the inverse transfer, from Market1501 to DukeMTMC, fails: we report a decrease of $2.7\%$ in Rank-1 accuracy, and of $0.3\%$ in mAP.

Our method is therefore effective for domain adaptation to the target dataset Market1501, but fails for the target dataset DukeMTMC. This result is found regardless of the source dataset. It seems that our method does not succeed as well in achieving the disentanglement of identity features on the images of the DukeMTMC dataset. Optimizing the learning parameters for this dataset, and especially the contributions of the various components of the cost function $\mathcal{L}_{Target}$, could perhaps help improve the disentanglement process, and therefore the performance of the ReID task. It should be noted also that the data available in the new domain is limited since it is only composed of person crops in which we select a subset of image pairs. Our approach would benefit from video data where many pairs could be built from trajectories.

We also report better performance for transfers from the MSMT17 dataset compared to the other source datasets, both for the performance in absolute value and with regard to the improvement compared to the source dataset. Our method delegates the learning of discriminative features to the source dataset, and its efficiency therefore strongly depends on the size of this dataset and the diversity of the images it contains. Since the MSMT17 dataset is larger and more diverse, it is expected to observe the best results for the transfers from this dataset.

The performance of transfers between the DukeMTMC databases and Market1501 is lower than the ECN \cite{zhongInvarianceMattersExemplar2019} and PDA-Net \cite{liCrossDatasetPersonReIdentification2019} methods. These methods are based more strongly on the specificities of the target dataset for the extraction of discriminating characteristics, by leveraging invariance properties or even using pose extraction. This explains their good results even when the source dataset is smaller. However, the performance of these methods on these transfers is comparable to our results for transfers from the MSMT17 dataset. The ECN \cite{zhongInvarianceMattersExemplar2019} and PDA-Net \cite {liCrossDatasetPersonReIdentification2019} methods have not been tested for transfers from the MSMT17 dataset.

\begin{figure}[htb]
\centering
\includegraphics[width=3.4in]{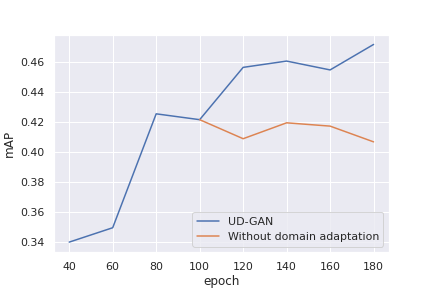}
\caption{Mean Average Precision on Market1501 dataset against training epochs on MSMT17 dataset, on two experimental settings : 1) using our framework and 2) without domain adaptation.}
\label{fig:TrainingCurve}
\end{figure}

\subsubsection{Training analysis}
For the transfer from MSMT17 to Market1501, we compare the domain transfer without domain adaptation and using our framework in Figure~\ref{fig:TrainingCurve}. We plotted the mAP on the target dataset Market1501 against the number of epochs of training on the MSMT17 dataset. Epochs are counted on the MSMT17 dataset : the first 100 training epochs correspond to stage 1, where only the ReID network is trained. The next epochs correspond to stage 3, in which we sample alternating batches from the source and target dataset. We compared with the direct transfer performance when the same amount of data from the source dataset is used during training. If the training on the source dataset MSMT17 continues without domain adaptation, the performance on the target dataset Market1501 quickly reaches a plateau, and even decreases as the ReID network is optimized for the source dataset. We conclude that the ReID network can lose general discriminative power by overfitting the source data distribution. Our method allows to pursue the training without overfitting using extra information from the target dataset. By doing so, the ReID network learns to extract relevant features in the target domain.

\subsubsection{Unsupervised pairs extraction}
We introduce a method for the unsupervised extraction of image pairs of same identity in the training set of the target domain at the beginning of stage 2. By simply considering the best match without additional processing, the pretrained ReID network on the MSMT17 dataset already achieves 89.1\% rank-1 accuracy in the unlabelled Market1501 training set. Note however that we allow here image matches captured by the same camera as the query image, contrarily to the evaluation protocol where matches have to be found within the images captured by the other cameras of the network. That explains why the accuracy is much higher than is the evaluation on the test set. This good result validates the choice of OSNet-AIN as a strong baseline for domain adaptive person ReID. 

After filtering out pairs that could be wrong according to the criteria described in the previous section, we achieve 96.0\% accuracy while keeping 78.8\% of the pairs (9602 pairs out of the possible 12185 pairs that we try to build with our pair extraction approach). We find similar performances for the precision of the pairs extraction during other transfers.


\section{Conclusion}

In this paper, we introduced a novel unsupervised disentanglement generative model (UD-GAN) for domain adaptive Person ReID. Our framework can be used with any backbone architecture, and on any unlabelled target domain without modification. We jointly train the backbone ReID network on the source labelled domain for discriminative identity features extraction, while conducting features disentanglement in the unlabelled target domain to adapt the ReID features to the new domain. We propose a method to reliably extract image pairs of the same identity in the target domain, that we use as inputs of our disentanglement generative framework. 

We conducted experiments on several standard ReID benchmarks for unsupervised Person ReID. We achieve state-of-the-art performance for mAP and Rank-1 accuracy for the transfer from MSMT17 to Market1501. However, our method does not always improve the performance of the baseline ReID network. Despite our efforts, transfers to the DukeMTMC dataset have been unsuccessful for now.

In the future, we plan on testing our framework with other backbone architectures. 


\section*{Acknowledgment}
We acknowledge the support of the Natural Sciences and Engineering Research Council of Canada (NSERC), [CRDPJ 528786 - 18], and the support of Arcturus Networks.


\bibliographystyle{IEEEtran}
\bibliography{Projet}

\end{document}